# Estimation of soil moisture in paddy field using Artificial Neural Networks

Chusnul Arif
Dept. of Civil and Environmental Engineering
Bogor Agricultural University
Bogor, Indonesia

Masaru Mizoguchi
Dept. of Global Agricultural Sciences
The University of Tokyo
Tokyo, Japan

Budi Indra Setiawan
Dept. of Civil and Environmental Engineering
Bogor Agricultural University
Bogor, Indonesia

Ryoichi Doi
Dept. of Global Agricultural Sciences
The University of Tokyo
Tokyo, Japan

***Abstract*—In paddy field, monitoring soil moisture is required for irrigation scheduling and water resource allocation, management and planning. The current study proposes an Artificial Neural Networks (ANN) model to estimate soil moisture in paddy field with limited meteorological data. Dynamic of ANN model was adopted to estimate soil moisture with the inputs of reference evapotranspiration ($ET_o$) and precipitation. $ET_o$ was firstly estimated using the maximum, average and minimum values of air temperature as the inputs of model. The models were performed under different weather conditions between the two paddy cultivation periods. Training process of model was carried out using the observation data in the first period, while validation process was conducted based on the observation data in the second period. Dynamic of ANN model estimated soil moisture with $R^2$ values of 0.80 and 0.73 for training and validation processes, respectively, indicated that tight linear correlations between observed and estimated values of soil moisture were observed. Thus, the ANN model reliably estimates soil moisture with limited meteorological data.**

## I. INTRODUCTION

In paddy field, soil moisture represents water availability for the plants and it is required for irrigation scheduling and water resource allocation, management and planning. Soil moisture variation affects the pattern of evapotranspiration, runoff and deep percolation in paddy field [1, 2, 3]. At the same time, soil moisture level is predominately influenced by water input through precipitation and irrigation.

However, hydrological data such as crop evapotranspiration, deep percolation, runoff and irrigation are often limited because acquisition of measurements in the field is costly, complicated, and time consuming. In addition, the detailed meteorological data required to determine crop evapotranspiration are not often available especially in developing countries. Therefore, estimation of soil moisture is needed using limited meteorological data.

Artificial Neural Networks (ANN) is suitable for use in dealing with complex system, such as agricultural system, than that mathematical method [4]. ANN has the capability to recognize and learn the underlying relations between input and output without explicit physical consideration [5]. The benefits of using ANN are its massively parallel distributed structure and the ability to learn and then generalize the problem [6]. In agricultural field, ANN has been applied to classify irrigation planning strategies [7] and to estimate subsurface wetting for drip irrigation [8].

The objective of this study was to estimate soil moisture from limited meteorological data using ANN model. Then, the proposed ANN model was validated by comparing observed and estimated values of soil moisture.

## II. MATERIALS AND METHODS

### A. Field Measurements

The field experiments were conducted in the experimental paddy field in the Nusantara Organics SRI Center (NOSC), Sukabumi, West Java, Indonesia during two paddy cultivation periods. The SRI center is located at 06º50’43” S and 106º48’20” E, at an altitude of 536 m above mean sea level. In both cultivation periods, meteorological data, consisted of air temperature and precipitation, were measured every 30 minutes. For validation model, soil moisture was measured using 5-TE sensor (Decagon Devices, Inc., USA) every 30 minutes.

TABLE I. PADDY CULTIVATIVATION PERIODS OF THE CURRENT STUDY

| Period | Planting date | Harvesting date | Season |
|---|---|---|---|
| First cultivation | 14 October 2010 | 8 February 2011 | Wet |
| Second cultivation | 20 August 2011 | 15 December 2011 | Dry - Wet |

All data were sent automatically to the server through quasi real-time monitoring [9, 10]. Daily maximum, average and minimum values of air temperature were used to estimate reference evapotranspiration ($ET_o$), then estimated $ET_o$ and precipitation were used to estimate soil moisture.

### B. Development of Artificial Neural Networks (ANN) models

Two ANN models were developed consisted of three layers, i.e. input, hidden and output layers, respectively (Fig.1).

The first model was developed to estimate $ET_o$ according to maximum, average and minimum values of air temperature (Fig.1a). $ET_o$ is a key variable in hydrological studies especially to quantitative knowledge of water supply, loss, and consumption in paddy field. Commonly, minimum meteorological data are required to determine $ET_o$ consisted of solar radiation and air temperature data using Hargreaves model [11]. However, solar radiation data are not often available, thus we only used air temperature data for $ET_o$ estimation. For validation, the output of model was validated by comparing to $ET_o$ derived by Hargreaves model [12].

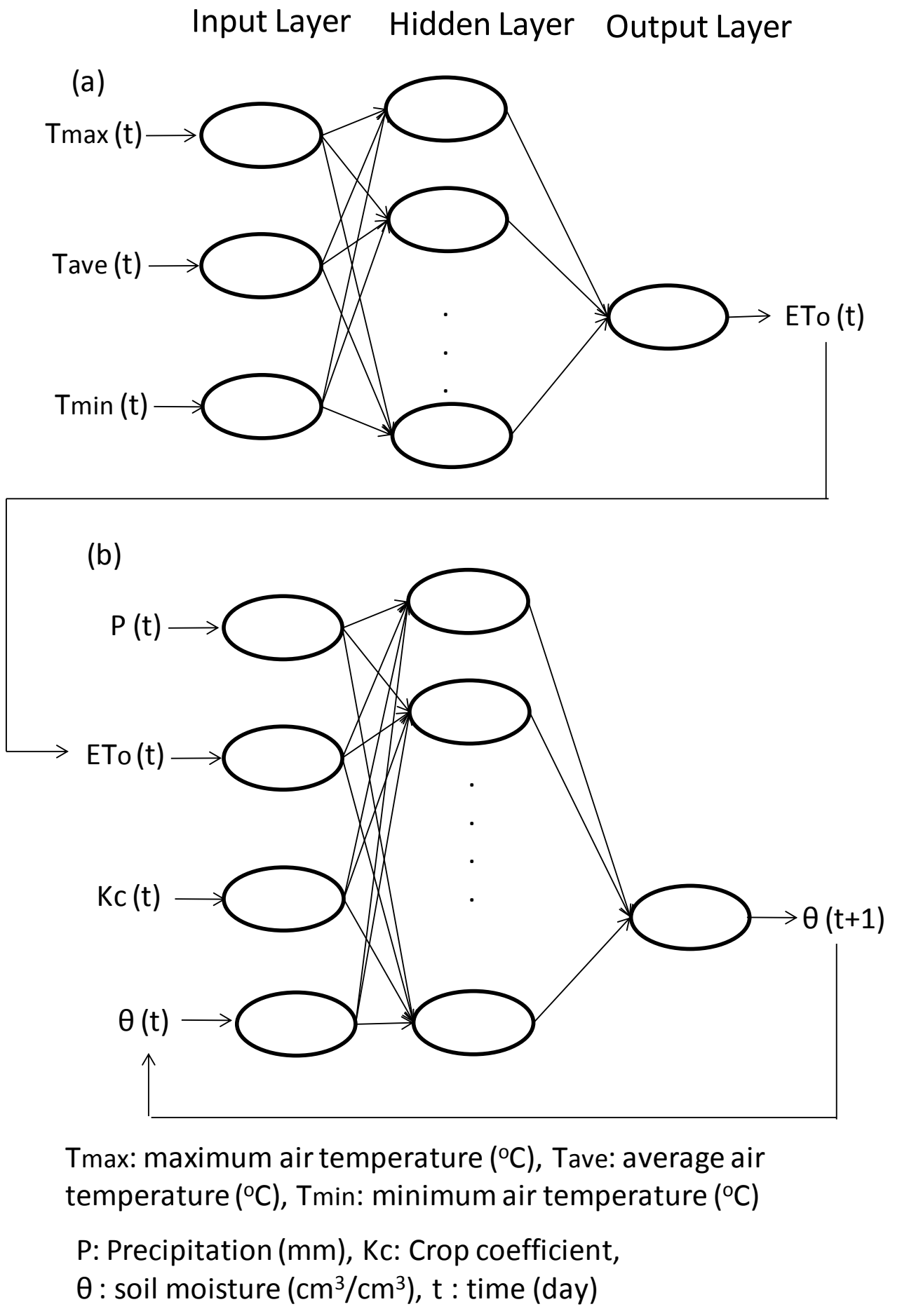

Figure 1. ANN models of the current study: a) ETo estimation, b) soil moisture estimation in paddy field.

Then, estimated $ET_o$ and precipitation data were used to estimate soil moisture (Fig. 1b). Soil moisture is usually determined by performing water balance analysis [1, 13]. The inflow of water balance consisted of precipitation and irrigation, while outflow consisted of crop evapotranspiration, runoff and deep percolation.

However, the variables such as irrigation, runoff and deep percolation were not available because the typical of measurements are complicated and expensive tools, heavy labor and time consumption. Meanwhile, crop evapotranspiration is derived by multiplying $ET_o$ to crop coefficient determined by the FAO procedure [14], thus crop coefficient was selected as the third inputs of ANN model (Fig.1b).

For this estimation, dynamic of ANN, similar to an auto-regressive moving average (ARMA) model procedure, was adopted by considering historical output data [15, 16]. All input parameters data were normalized between 0 and 1 by using fixed minimum and maximum values. Normalization of data is important to avoid larger numbers from overriding smaller ones and to prevent premature saturation of hidden nodes, which impedes the learning process [5]. For hidden layer, total nodes of 8 were selected as moderate number to prevent excessively time-consuming in training process and to avoid incapability in differentiating between complex patterns due to few number of node in hidden layer.

Back propagation was selected as the learning method, which is composed of two phases; first, propagation (forward and backward propagation), and second, weight update. A sigmoid function was selected as the activation by the following equations:

$$f(y) = \frac{1}{1 + e^{-gy}} \tag{1}$$

$$y = \sum_{i=0}^{n} x_i w_i \tag{2}$$

where $x_i$, $w_i$, n, g are the inputs, weights, number of inputs and gain parameter, respectively.

The gain parameters are usually set to 1.0 and not changed by the learning rule. However, this fixed value probably caused local minima problem. Therefore, the gain parameter should be adjusted according to the degree of approximation to the desired output of the output layer [17]. Here, gain parameter is adjusted based on the following condition:

$$g = \begin{cases} \frac{1}{Ap} if\ \text{Ap} > 1.0 \\ 1.0\ if \text{Ap} \le 1.0 \end{cases} \tag{3}$$

$$Ap = 2e_p \tag{4}$$

$$e_p = \max\left|t_p - o_p\right| \tag{5}$$

where Ap is degree of the output layer, $e_p$ is error for pattern *p*, $t_p$ and $o_p$ are observed and estimated output for pattern *p*, respectively.

The observed data were divided into two data sets. The observed data from first paddy cultivation period were used for training process, while data from the second period were used for validation process. This kind of validation is called as a cross validation method [18, 19]. Then, developed ANN model was evaluated by comparing observed and estimated values of soil moisture using the indicator of coefficient determination ($R^2$). The value of $R^2$ ranged from 0.0 to 1.0 with higher values indicating better agreement.

## III. Results And Discussion

### A. Weather conditions during paddy cultivation periods

Meteorological conditions in the first and second paddy cultivation periods are shown in Fig. 2-3. In the first paddy cultivation period, the meteorological parameters were characterized by low air temperature and high precipitation compared to the second period. The monthly average air temperature changed during in the end of 2010 and 2011, where its value was highest on November 2010 for the first period, and then it occurred on December 2011 for the second period. The highest average air temperature values were 23.87°C and 24.49°C for first and second periods, respectively. The same situation occurred to precipitation in which highest total precipitation was 400.8 mm on December 2010 for the first period and its value was 369.6 mm on November 2011 for the second period.

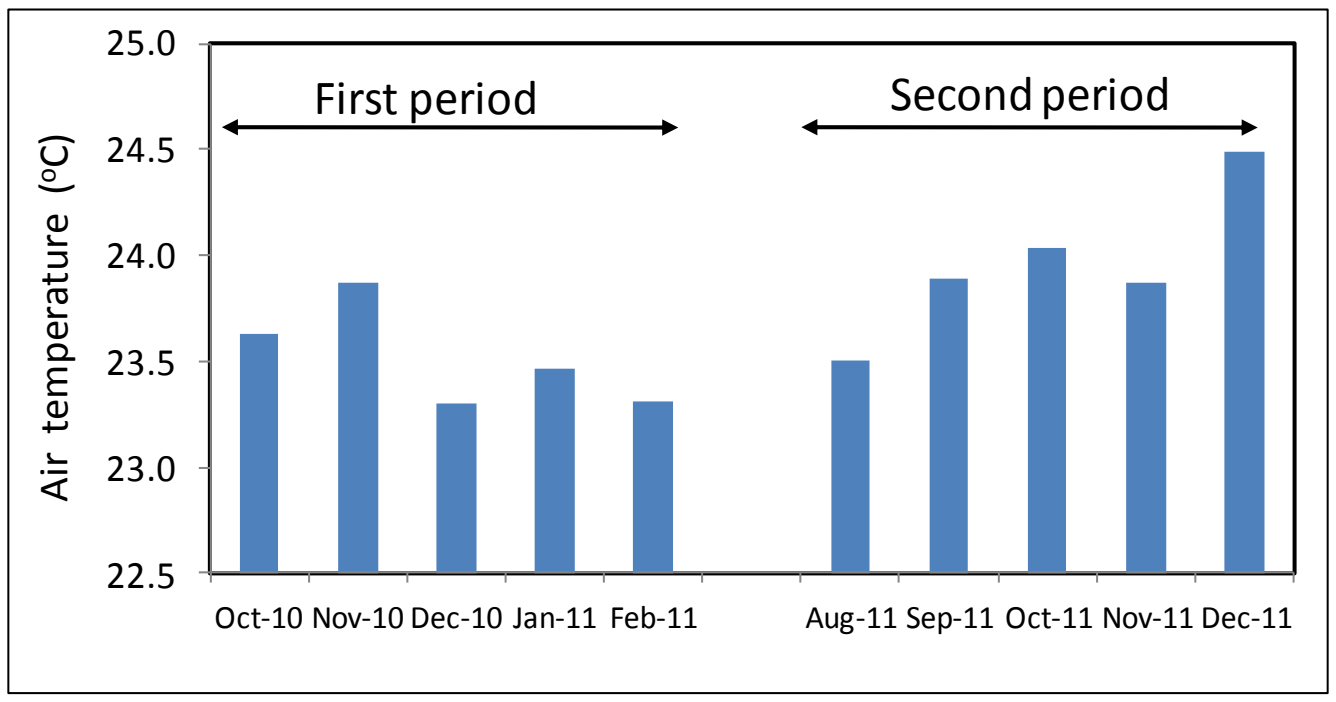

Figure 2. Monthly average air temperature during cultivation periods.

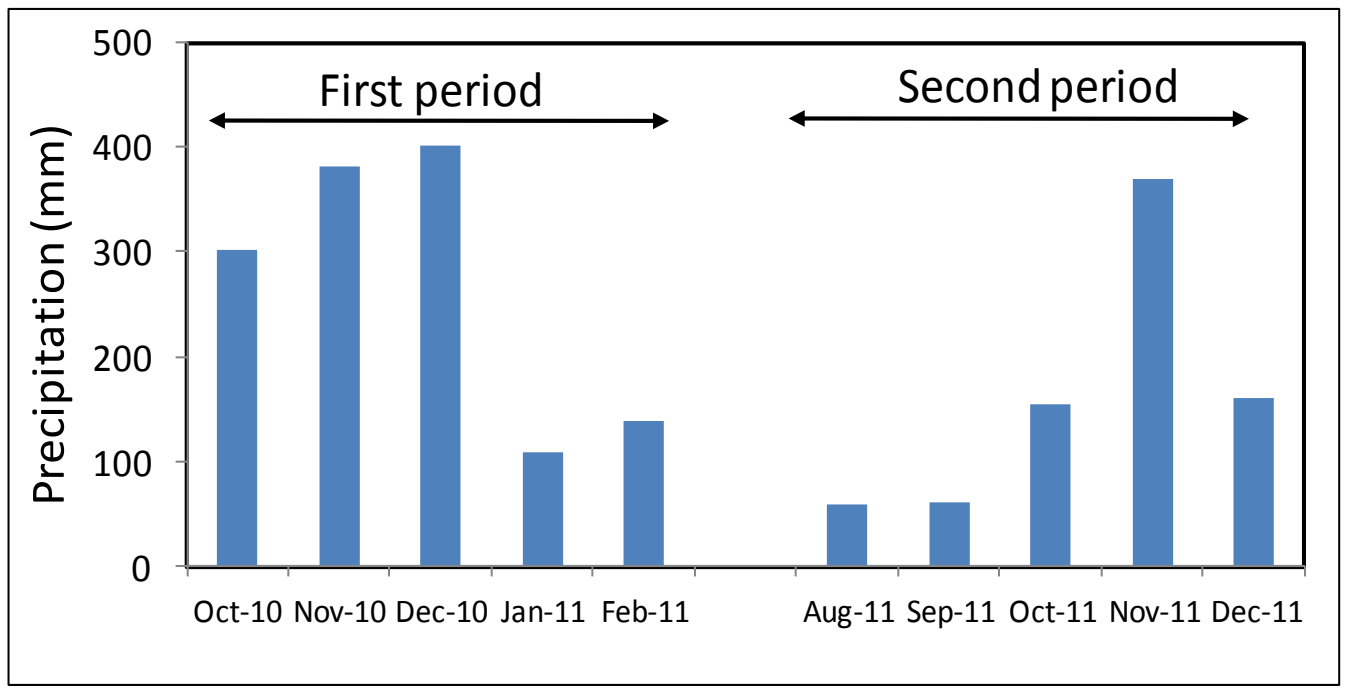

Figure 3. Total precipitation during cultivation periods.

### B. Estimation of reference evapotranspiration

Training process should be carried out firstly by ANN model to learn the pattern of observation data between the input and the output. In this process, a thousand iterations were performed to minimize the error of estimation. As the result, ANN model estimated $ET_o$ with $R^2$ of 0.96 as shown in Fig. 4a. Then, the weights, results of training process, were used to estimate $ET_o$ using the second period data with the result as shown in Fig. 4b. Underestimation was occurred when $ET_o$ value was higher than 4.5 mm.

However, with $R^2$ of 0.95, the estimated $ET_o$ showed good agreement to the Hargreaves model. Therefore, ANN model can be used to estimate $ET_o$ using the inputs of maximum, average and minimum values of air temperature.

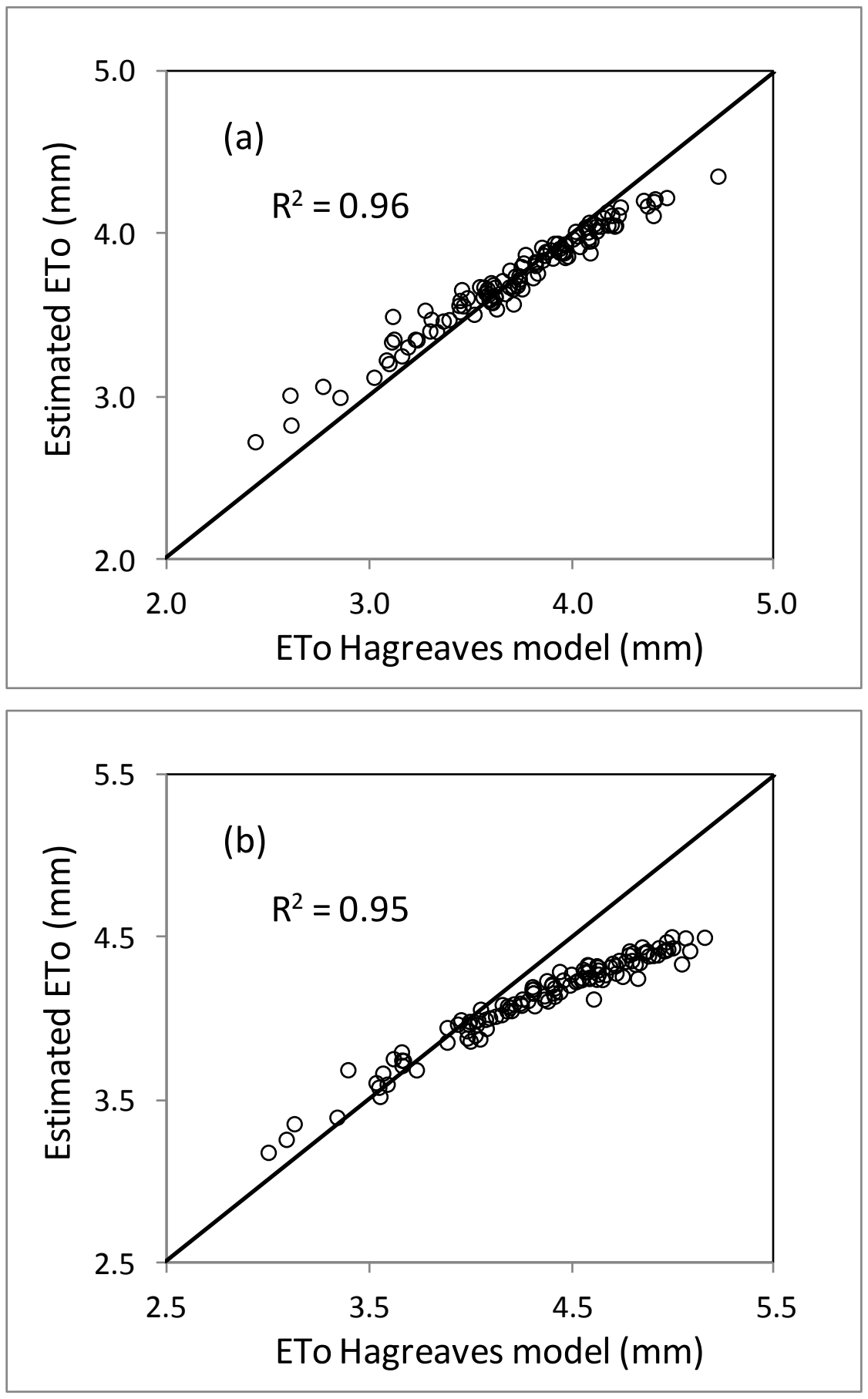

Figure 4. Model evaluation of ANN model for ETo estimation: a) training process, b) validation process.

### C. Estimation of soil moisture

Dynamic of ANN model estimated soil moisture with $R^2$ of 0.80 and 0.73 for both training and validation processes, respectively (Fig. 5). In the training process, tight linear correlations between observed and estimated values of soil moisture were observed (Fig. 5a), thus more than 80% of the changes in observed soil moisture were well described by the model. Therefore, the weights, representation of relationship between the input and the output model, can be used to estimate soil moisture in paddy field with limited meteorological data.

Fig. 5b shows the correlation between observed and estimated values of soil moisture in the second cultivation period as the result of validation process. ANN model estimated soil moisture with $R^2$ values of greater than 0.72 indicate the model's performance.

Thus, tight linear correlations between observed and estimated values of soil moisture by the ANN model (Fig. 5b). Accordingly, the estimation results in this study showed good agreement to the observed data, thus the proposed method can be accepted as suggested by the previous study [1].

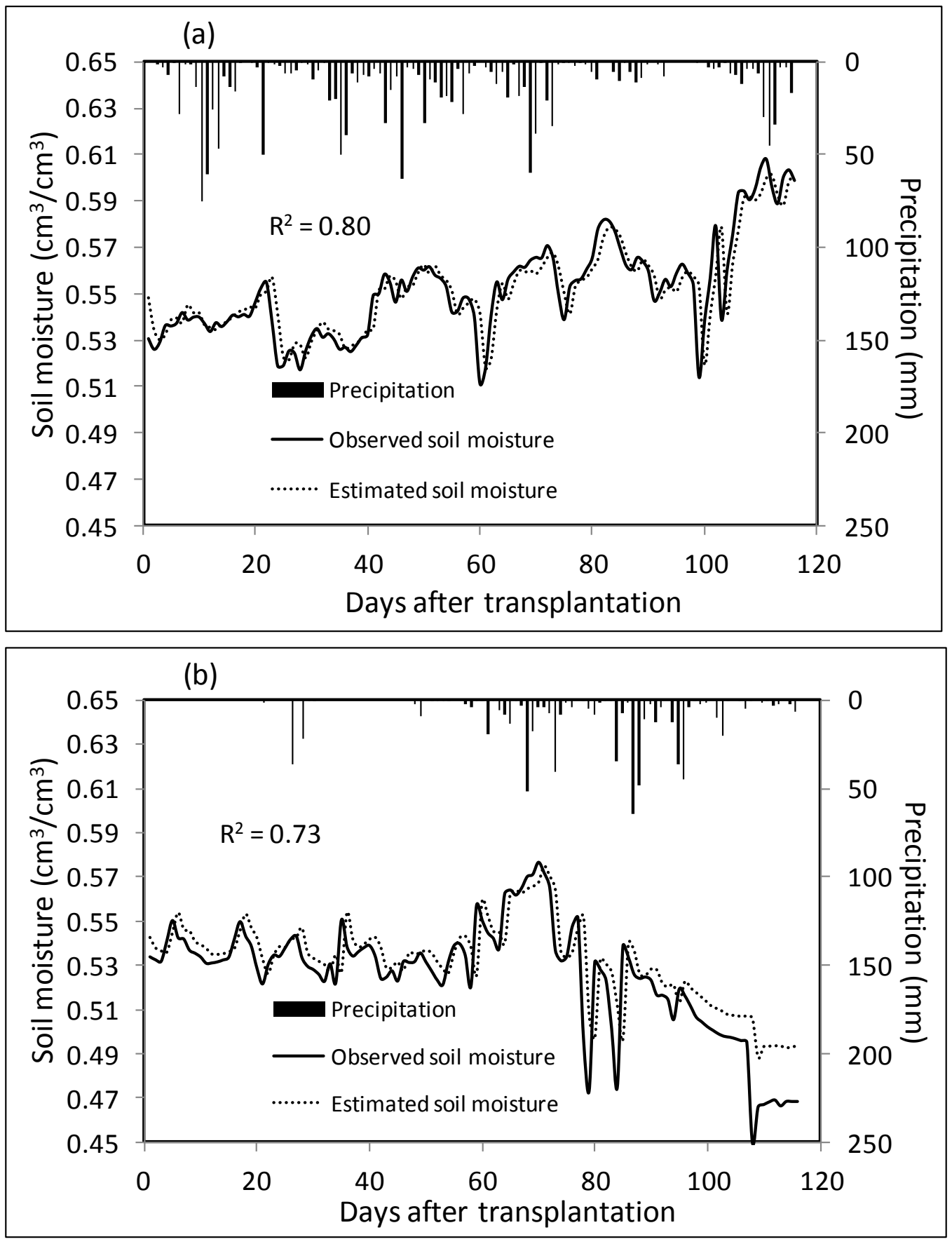

Figure 5. Observed and estimated soil moisture during cultivation periods: a) first period, b) second period.

## IV. CONCLUSIONS

The current study proposed Artificial Neural Networks (ANN) models to estimate soil moisture in paddy field with limited meteorological data. Two ANN models were proposed, first model was developed to estimate reference evapotranspiration ($ET_o$) using the maximum, average and minimum values of air temperature as the inputs. Then, estimated $ET_o$ and precipitation were used to estimate soil moisture by adopting dynamic of ANN in the second model.

The proposed models were performed under different weather conditions between the two paddy cultivation periods. In the first model, $ET_o$ was estimated accurately with $R^2$ of 0.96 and 0.95 for training and validation processes, respectively. Then, the second model estimated soil moisture with $R^2$ values of greater than 0.72 for training and validation processes suggested that the ANN model was reliably to estimate soil moisture in paddy field with limited meteorological data and without complicated and expensive tools, heavy labor and time consumption.

## ACKNOWLEDGMENT

The authors are grateful to the Directorate of Higher Education, Ministry of National Education, Republic of Indonesia for generous financial support through grant of International Research Collaboration and Scientific Publication. Also, the authors acknowledge the financial support by the Japan Society for the Promotion of Science.

AUTHORS PROFILE

**Chusnul Arif**, lecturer at Department of Civil and Environmental Engineering, Bogor Agricultural University, Indonesia. He is interested study on the application of Artificial Intelligence such as Artifical Neural Networks and Genetic Algorithms in agricultural field particularly in paddy field. Since April 2010, he is pursuing PhD degree at Department of Global Agricultural Sciences, the Univeristy of Tokyo with the main topic is optimization of water management in paddy field. This paper is part of his PhD dissertation.

**Budi Indra Setiawan**, Professor of soil physic and hydrology at Department of Civil and Environmental Engineering, Bogor Agricultural University, Indonesia. His common interest is to develop theory, model and technology which have direct benefits for the improvement of water resources management in agricultures as well as aquacultures.

**Masaru Mizoguchi**, Professor of International agro-informatics at Department of Global Agricultural Sciences, the University of Tokyo, Japan. He is developing quasi real-time monitoring of farmland for any purpose and study, such as climate change and plant growth model.

**Ryoichi Doi**, Assistant professor at Department of Global Agricultural Sciences, the University of Tokyo, Japan, has been working interdisciplinary in areas of appropriate technology, environment and agriculture related to socio-economic development.